\documentclass[letterpaper, 10pt, conference]{ieeeconf}  

\IEEEoverridecommandlockouts                              

\usepackage{graphicx}
\usepackage{longtable}
\usepackage{url}
\usepackage{graphicx}
\usepackage{float}
\usepackage{array}
\usepackage{pgfgantt}
\usepackage{times}
\usepackage{subcaption}
\usepackage{caption}
\usepackage{pdflscape} 
\usepackage{ragged2e} 
\usepackage{mathptmx}
\usepackage{anyfontsize}
\usepackage{t1enc}
\usepackage{setspace}
\usepackage[many]{tcolorbox}
\usepackage{lipsum}
\usepackage[colorlinks, allcolors=blue]{hyperref}
\usepackage{verbatim}
\usepackage{cite}
\usepackage{todonotes}
\usepackage{pgfplots}
\usepackage{dblfloatfix}
\usepackage{xcolor}
\usepackage{pifont}

\usepackage{tikz}

\definecolor{royalblue}{rgb}{0.25, 0.41, 0.88}
\definecolor{tifanny_blue}{HTML}{00B3B9}
\definecolor{field_drab}{HTML}{5E4D00}
\definecolor{arctic_lime}{HTML}{D3FF11}
\definecolor{spring_frost}{HTML}{84FF42}
\definecolor{sky_blue}{HTML}{36CDFF}
\definecolor{orange}{HTML}{FF7F00}
\definecolor{azul}{HTML}{107896}
\definecolor{naranja}{HTML}{C2571A}
\definecolor{rojo}{HTML}{EE4B2B}
\definecolor{amarillo}{HTML}{BCA136}
\definecolor{verde}{HTML}{4CBB17}
\definecolor{gris}{HTML}{909090}
\definecolor{rosa}{HTML}{F9A7B0}
\definecolor{amarillochillon}{HTML}{FBB117}
\definecolor{olive}{rgb}{0.42, 0.56, 0.14}
\definecolor{copper}{rgb}{0.72, 0.45, 0.2}

\definecolor{reviewed_color}{rgb}{0, 0, 0.0}

\newcolumntype{P}[1]{>{\centering\arraybackslash}p{#1}}

\newcommand{\cmark}{\textcolor{verde}{\textbf{\ding{52}}}}%
\newcommand{\xmark}{\textcolor{rojo}{\small{\ding{55}}}}%

\usetikzlibrary{mindmap, matrix, shapes.geometric, arrows, calc, patterns}
\tikzstyle{startstop} = [rectangle, rounded corners, minimum width=3cm, minimum height=1cm,text centered, draw=black, fill=red!30]
\tikzstyle{io} = [trapezium, trapezium left angle=70, trapezium right angle=110, minimum height=1cm, text centered, draw=black, fill=blue!30]
\tikzstyle{process} = [rectangle, minimum width=3cm, minimum height=1cm, text centered, draw=black, fill=orange!30]
\tikzstyle{decision} = [diamond, minimum width=3cm, minimum height=1cm, text centered, draw=black, fill=green!30]
\tikzstyle{arrow} = [thick,->,>=stealth]
\tikzstyle{darrow} = [thick,<->,>=stealth]
\tikzstyle{warrow} = [thick,->,>=stealth]

\newcommand{\writetext}[1]{\todo[inline, color=red!40]{WRITE}}

\newcommand{\mycomment}[1]{}

\title{\LARGE \bf
Fields2Cover: An open-source coverage path planning library for unmanned agricultural vehicles
}

\author{Gonzalo Mier$^{1}$ \and João Valente$^{2}$ \and Sytze de Bruin$^{3}$
\thanks{$^{1}$Gonzalo Mier with the Laboratory of Geo-Information Science and Remote Sensing, Wageningen University \& Research, Wageningen, The Netherlands {\tt\small gonzalo.miermunoz@wur.nl}}
\thanks{$^{2}$João Valente with the Information Technology Group, Wageningen University \& Research, Wageningen, The Netherlands}
\thanks{$^{3}$Sytze de Bruin with the Laboratory of Geo-Information Science and Remote Sensing, Wageningen University \& Research, Wageningen, The Netherlands}%
}

\begin{document}

\maketitle
\thispagestyle{empty}
\pagestyle{empty}

\begin{abstract}

This paper describes Fields2Cover\footnote{https://github.com/Fields2Cover/Fields2Cover}, a novel open source library for coverage path planning (CPP) for agricultural vehicles. While there are several CPP solutions nowadays, there have been limited efforts to unify them into an open source library and provide benchmarking tools to compare their performance. Fields2Cover provides a framework for planning coverage paths, developing novel techniques, and benchmarking state-of-the-art algorithms. The library features a modular and extensible architecture that supports various vehicles and can be used for a variety of applications, including farms. Its core modules are: a headland generator, a swath generator, a route planner and a path planner. An interface to the Robot Operating System (ROS) is also supplied as an add-on. In this paper, the functionalities of the library for planning a coverage path in agriculture are demonstrated using 8 state-of-the-art methods and 7 objective functions in simulation and field experiments.

Keywords -- Agricultural Automation, Software Architecture for Robotic and Automation, Field Robots
\end{abstract}

\section{Introduction}

In developed countries, there is a shortage of skilled workers to operate agricultural machinery \cite{christiaensen2020future}. This shortage can be alleviated with the development of autonomous machinery. Unlike manually operated machinery, autonomous vehicle operations need meticulous planning beforehand. The problem of determining a path to cover a field is known as coverage path planning (CPP). CPP is of high importance for cleaning \cite{bormann2018indoor}, surveillance robots \cite{jensen2020near}, lawn mowers \cite{hameed2017coverage}, and agricultural vehicles \cite{oksanen2009coverage}, where it has been addressed in several works. 

Whilst there have been many efforts, most of the (partial) CPP solutions have not been released as open-source software thus hindering more rapid advances in CPP by the scientific community. The packages shown in Table \ref{table:related_works} are the only open-source software to the best of our knowledge. Note that the software packages listed in Table \ref{table:related_works} solve the CPP problem partially, but require several modifications in order to be customized to different unmanned vehicles and applications.

This paper aims to fill the above mentioned gap by proposing and releasing to the community an open-source CPP library for field coverage. The library was designed focusing in four modules that are the core of CPP solutions: a headland generator, a swath generator, a route planner, and a path planner. Each module includes at least one state-of-the-art method and one objective function. The library currently only supports convex fields \textcolor{reviewed_color}{on arable farmland}. Regardless, there is an urgent need for an open source software solution to fill the existing gap in the CPP problem in agriculture. The ultimate goal of the library is to ease the state of-the-art algorithm benchmark and to accelerate CPP research and application.

\begin{table*}[]
\centering
\caption{Comparison between coverage path planning open-source software solutions. \textcolor{reviewed_color}{Repositories (rows) are compared in terms of (1) available documentation (Docs); (2) Computation of exact solutions rather than using a  discretizing grids (No grid); (3) Support for non-holonomous vehicles in turns; (4) The option to reserve maneuvering space at the field boundaries (Headlands support); (5) The possibility to modify the objective function; (6) Applicability for agricultural ground robots.}}
\scalebox{0.85}{
\begin{tabular}{| c | P{0.7cm} | P{1.1cm}  | P{1.5cm} | P{1.5cm} | P{1.7cm} | P{1.8cm} |} 
 \hline
    Package name & \textit{\small{Docs}} & \textit{\small{No grid used}} & \textit{\small{Non-holonomous}} & \textit{\small{Headlands support}}  &  \textit{\small{Customizable objective function}} & \textit{\small{ Terrestrial agricultural vehicles}}\\ [0.5ex]
 \hline\hline

 \small{RJJxp/CoveragePlanning \cite{RJJxpCoveragePlanninggithub}}  & \xmark & \cmark & \xmark & \xmark & \xmark & \xmark \\ \hline
 \small{Nobleo/full\_coverage\_path\_planner \cite{Nobleogithub}}  & \cmark & \xmark & \cmark& \xmark & \xmark & \xmark \\ \hline
  \small{Ipa320/ipa\_coverage\_planning \cite{ipa320ipacoverageplanninggithub, bormann2016room, bormann2018indoor}} & \cmark & \cmark & \cmark & \xmark & \xmark & \xmark \\ \hline
 \small{Ethz-asl/polygon\_coverage\_planning \cite{Ethzaslgithub,Ethzaslgithubbahnemann2021revisiting}}  & \cmark & \cmark & \xmark & \xmark & \xmark & \xmark \\ \hline
 \small{Irvingvasquez/ocpp \cite{irvingvasquezocppgithub, irvingvasquezocppgithubgomez2017optimal}}  & \xmark & \cmark & \xmark & \xmark & \xmark & \xmark \\ \hline
  \small{Greenzie/boustrophedon\_planner \cite{Greenziegithub}} & \xmark & \cmark& \cmark & \cmark & \xmark & \xmark \\ \hline
 \small{Ipiano/coverage-planning \cite{IpianoCoveragePlanninggithub, driscoll2011complete}} & \cmark & \cmark & \xmark & \xmark & \cmark & \xmark \\ \hline
 \textbf{Fields2Cover}  & \cmark  & \cmark & \cmark & \cmark & \cmark & \cmark \\ \hline 
\end{tabular}
}
\label{table:related_works}
\end{table*}

\subsection{Related work}

Owing to the non-holonomous nature of agricultural vehicles, a region of the field known as headlands must be reserved for turning the vehicle. The most basic approach is to allocate a constant width area around the field. This strategy allocates a large amount of space to a poor yield area. Depending on how the swaths are arranged, some headland areas are parallel to the swaths and hence they are not needed for turning. By only constructing headlands along the field edges where turns are made, the area reserved for them can be minimized \cite{oksanen2009coverage, jin2009optimal}.

Swaths are generated in the inner field, which is the remaining region after subtracting the headlands. In two-dimensional planar fields, a reference line can be applied as a guide for the generation of swaths, where each parallel creates a swath \cite{jin2009optimal, oksanen2009coverage, debruin2014gaos}. This line can be chosen for convenience or by an algorithm such as brute force or a meta-heuristic. Oksanen \cite{oksanen2009coverage} describes a driving angle search strategy that requires fewer iterations than brute force search but it does not guarantee finding the global minimum. Objective functions such as the number of turns  or the sum of swath lengths are used to determine optimality in swath generation \cite{jin2009optimal}.

The distance \cite{jin2009optimal} and time \cite{meuth2008divide} required to cover the field are affected by the order of the swaths. A route is the sequence of the swaths to cover. The Boustrophedon order, which travels the swaths sequentially from one side of the field to the other, and the snake order, which skips one swath at each turn and returns through the uncovered swaths, are popular preset routing patterns \cite{zhou2015quantifying}. Objective functions such as distance, number of rotations, or time necessary to traverse the field \cite{jin2009optimal, meuth2008divide} are minimized by finding the optimal route through meta-heuristics \cite{spekken2016planning}. 

A path is composed of the swaths of a route connected by turns, forming a continuous line along which the vehicle will drive. Dubins' \cite{dubins1957curves} or Reeds-Shepp's \cite{reeds1990optimal} curves are turns that minimize the path length of the turns. These curves are made by either curve segments or straight lines. The main problem is that there is an instantaneous change of curvature at the transition point between two segments. Techniques such as numerical integrators \cite{backman2015smooth} or clothoids \cite{sabelhaus2013using} are employed to smooth the turn to avoid the curvature discontinuity. Furthermore, to navigate from a swath to the headlands, turns such as non-uniform rational B-spline (NURBS) curves \cite{hoffmann2022weight} can be adopted.

CPP problems are composed of numerous sub-problems, several of which have received special attention in literature. For example, Spekken \cite{spekken2016planning} presents an approach for calculating the coverage path in undulating terrain that however does not consider turns between rows or headland creation. Nilsson \cite{nilsson2020method} and Nørremark \cite{norremark2022field} divide the CPP problem into two major modules: Field Partitioning/Representation, where the distribution of headlands and swaths in the field is set up, and Route Planning, which determines the optimal order of travelling the swaths within sub-fields. In the latter framework, each module has more than one function, increasing the complexity of comparing multiple variations of the module.

\subsection{Existing open-source software}

There have been web applications, such as GAOS \cite{debruin2014gaos}, that allowed farmers to design or adapt coverage paths with a user-friendly interface. Many of such web applications, despite being a great help to the farming community, have been developed in collaboration with companies, restricting the possibility to release the code to the public domain.

The currently existing open source CPP repositories are listed in Table \ref{table:related_works}. Although seven other projects were found, none of them can be adopted for farming purposes with ground robots. As mentioned above, ground robots in agriculture are generally non-holonomous, so turning maneuvers must be planned to move from one swath to another. Unfortunately, some packages \cite{RJJxpCoveragePlanninggithub, Ethzaslgithub,irvingvasquezocppgithub, IpianoCoveragePlanninggithub} only compute the route to cover a region. These packages are designed for quadrotors \cite{Ethzaslgithub, irvingvasquezocppgithub} or for indoor robots \cite{RJJxpCoveragePlanninggithub}. However, the code needs to be modified to support path generation for non-holonomous robots. A special case of CPP is the Nobleo package \cite{Nobleogithub} which, although the vehicle used is non-holonomous, uses a grid to define the nodes that should be covered at least once. In agriculture it is important to reduce the damage caused by the wheels of the vehicle, so it is not recommended to cover the same swath several times \cite{Nobleogithub} or to cross through the main field \cite{RJJxpCoveragePlanninggithub, Ethzaslgithub, ipa320ipacoverageplanninggithub}. On the other hand, Greenzie \cite{Greenziegithub}, which was developed for lawn mowers, is the only package that supports headlands, along with Fields2Cover. Unlike arable farming, mowers are constrained to avoid repeated tracks for field traffic, thus the coverage path is created with random sweep angles. For this reason, Greenzie does neither provide an optimizer nor an objective function for planning the swaths. In contrast, Ipiano \cite{IpianoCoveragePlanninggithub}  provides an interface to change the objective function used by its optimizer, but here no headland support is offered. Fields2Cover is the only software solution that provides algorithms to create a coverage path for terrestrial agricultural robots, including optimizers and objective functions to generate the best path, headland support and turn planning.

\subsection{Contributions}

\textcolor{reviewed_color}{The main contributions of this paper and the Fields2Cover library are:}
\begin{enumerate}
    \item A publicly-available library (Fields2cover) providing connectable modules to address CPP problems with unmanned agricultural vehicles. Those modules can be effortlessly customized for other CPP problems.
    \item Benchmark tools for quantitative comparison between the CPP algorithms and approaches.   
    \item A quantitative comparison using 38 convex fields between eight state-of-the-art CPP approaches/methods and seven objective functions.
    \item Experiments with a commercial unmanned agricultural vehicle demonstrating Fields2Cover's capability to provide real-world solutions.
    \item \textcolor{reviewed_color}{Contribute to building a research community by providing an accessible platform for discussing novel ideas, solving issues and sharing new variants of the problem.}
\end{enumerate}

\section{Fields2Cover}
\label{sec:fields2cover}

\begin{figure}[]
\centering
\includegraphics[width=1\linewidth]{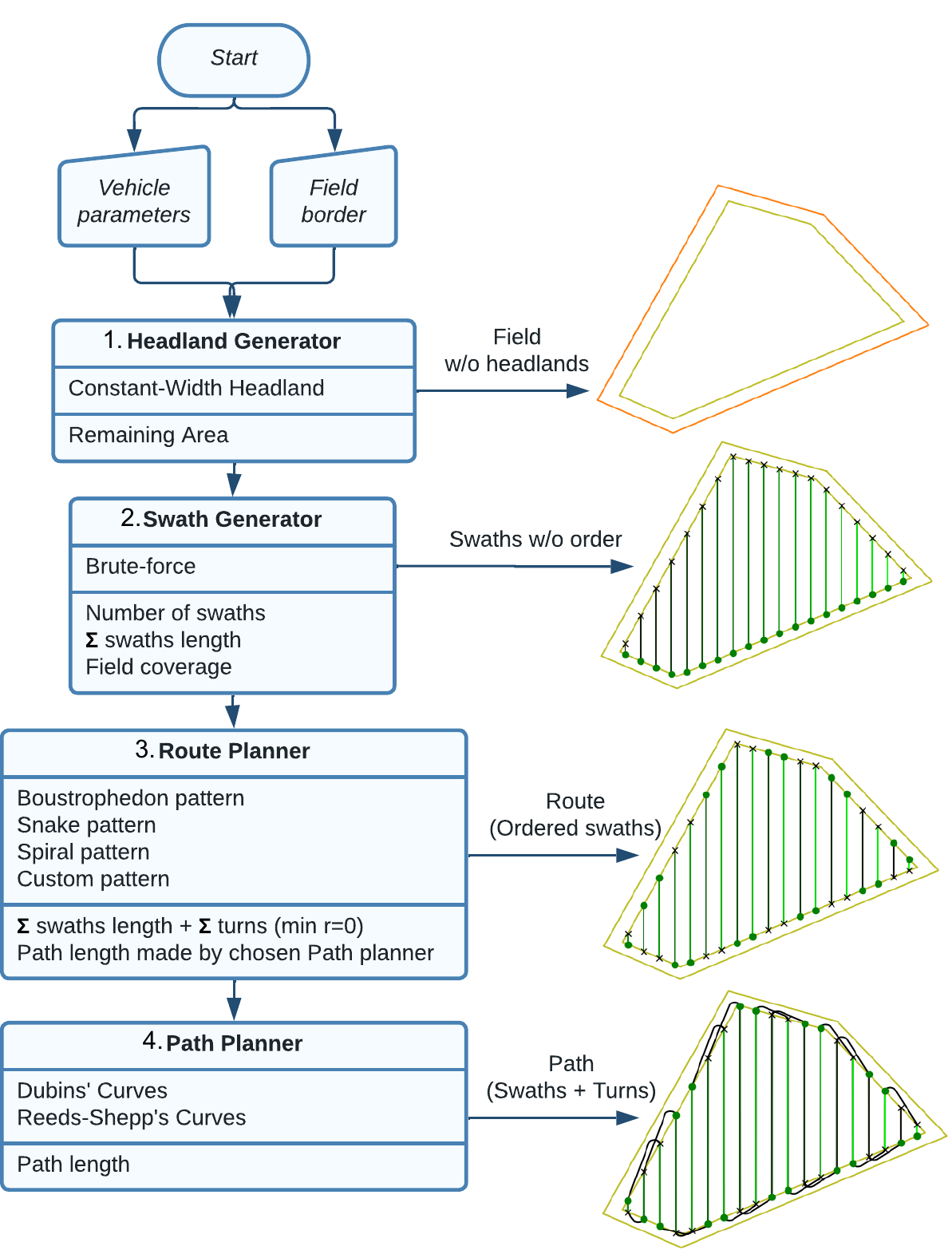}
\caption{Diagram of the Fields2Cover library. The library contains four core modules: 1) Headland Generator, 2) Swath Generator, 3) Route Planner and 4) Path Planner. Each of the modules is represented as a box with 3 slots: the name of the module, the methods implemented, and the objective functions available. The output of a module is the input of the next module.}
\label{fig:diagram}
\end{figure}

Fields2Cover is designed in four modules (Fig. \ref{fig:diagram}): 1) Headland Generator, 2) Swath Generator, 3) Route Planner and 4) Path Planner. The inputs of the CPP problem are the shape of the field and the vehicle specifications, while the output is the coverage path of the field. Methods from the same module can be used interchangeably to compare their solutions independently from the rest of the CPP problem.

\subsection{Headland Generator module}

The Headland Generator module currently implements a single method that buffers the border of the field in inward direction by a custom constant width (see Module 1 in Fig. \ref{fig:diagram}). The objective function of this module is the area of the remaining field after removing the headlands.

\begin{equation}
    A_r = \frac{A_{\bar{hl}}}{A_{f}}
\end{equation}

\textcolor{reviewed_color}{where $A_{\bar{hl}}$ is the area of the field without headlands (mainland), $A_{f}$ the area of the original field, and $A_r$ is the ratio of mainland to the original field.}

\subsection{Swath Generator module}

The inner field (i.e., excluding the headlands) is the input of the Swath Generator module (see Module 2 in Fig. \ref{fig:diagram}). 
This region is divided into parallel swaths matching the operating width. In the current version, the library only supports parallel non-overlapping swaths. Fields2Cover has a brute force algorithm to find the optimal sweep angle by trying discretized angles using a given step size. If the computer running the library supports multiple threads, several sweep angles are tried in parallel \cite{TBBsoftware}.

This module currently implements 3 objective functions:
\begin{itemize}
    \item Minimize the Number of Swaths. This objective function depends on the shape and the area of the field, and the width of the robot. The number of swaths is limited by the equation:
    \begin{equation}
      0 \leq \# S_{\alpha} \leq \frac{A_{\bar{hl}}}{R_w},
    \end{equation}
    where $\# S_{\alpha}$ is the number of swaths for a given sweep angle $\alpha$, $A_{\bar{hl}}$ is the area of the field without headland, and $R_w$ is the operational width of the robot. The shape of the field that maximizes the minimun number of swaths is the square field, which results in:
    \begin{equation}
      min_\alpha\ \# S_{\alpha}^{\tiny\diamondsuit} \simeq \frac{\sqrt{A_{\bar{hl}}}}{R_w},
    \end{equation}
    where $\# S_{\alpha}^{\tiny\diamondsuit}$ is number of swaths in a square field with a given  sweep angle $\alpha$. Therefore, the optimal value of this objective function is less than the square root of the area of the field.

\item Maximize the Field Coverage:
\begin{equation}
    A_{cov} = \frac{A_{\bar{hl}} \cap \{\cup_i\ S^i \}) }{A_{\bar{hl}}},
\end{equation}

where $A_{cov}$ is the \textcolor{reviewed_color}{fraction} of area covered, $A_{\bar{hl}}$ is the field without headlands, $S^i$ is the $i^{th}$ swath, $\cap$ is the intersection operator, and $ (\cup_i\ S^i)$ is the union of all the swaths.

\item Minimize the Swaths Length:
\begin{equation}
   \sum_i^{N} length(S^i) = \sum_{i}^{N} \sum_{j}^{S^i_p - 1} ||S^i_{p=j+1} - S^i_{p=j}||_2,
\end{equation}
where $\sum_i length(S^i)$ is the sum of the length of the swaths, $N$ is the number of swaths, $S^i_p$ is the number of points that the $i^{th}$ swath has, $^i_{p=j}$ is the $j^{th}$ point of the $i^{th}$ swath, and $||x||_2$ is the Euclidian norm.
\end{itemize}

\subsection{Route Planner module}

The Route Planner module uses the swaths created earlier to produce the route (see Module 3 in Fig. \ref{fig:diagram}). Fields2Cover contains several predefined route patterns, which include the boustrophedon pattern, the snake pattern, the spiral pattern and a custom pattern. The Boustrophedon pattern covers the swaths sequentially, and the Snake pattern skips one swath each time to traverse the field in one direction and returns through covering the uncovered swaths. The Spiral pattern is a variation of the Snake pattern, that sorts the swaths in clusters of a fixed size with the snake pattern. The custom pattern requires specification of the swath order by the user. To compare different routes, the library provides as objective function the length of the path generated by the Path Planning module. \textcolor{reviewed_color}{It also computes the path length with in-place turns, which correspond to zero  turning radius. The path length of in-place turns is computed as :}
\begin{equation}
   \textcolor{reviewed_color}{ L_0 = \sum_{i=1}^{N} length(S^i) + \sum_{i=2}^{N} ||S^i_{p=1} - S^{i-1}_{p=M}||_2}
\end{equation}

\textcolor{reviewed_color}{where $L_0$ is the path length with in-place turns, $\sum_{i=1}^{N} length(S^i)$ is the sum of the lengths of the swaths, $N$ is the number of swaths, $S^i_{p=1}$ is the first point of the $i^{th}$ swath, $S^i_{p=M}$ is the last point of the $i^{th}$ swath, and $||x||_2$ is the Euclidian norm.}

\subsection{Path Planner module}

The inputs of the Path Planner module (see Module 4 in Fig. \ref{fig:diagram}) are the route (sorted swaths) and the vehicle parameters.
Once the route is known, the turns to complete the path are computed. In the current version of the library, the path planner applies the same type of curves for all the headland turns. Fields2Cover currently supports straight curves, the Dubins' curves\cite{dubins1957curves} and the Reeds-Shepp's\cite{reeds1990optimal} curves, using the path length as the single objective function.

\subsection{ROS wrapper}
\label{subsec:ros}

Although the Fields2Cover library does not depend on ROS, an interface with \textcolor{reviewed_color}{ROS1 and ROS2} is provided as an add-on. The \textcolor{reviewed_color}{fields2cover\_ros}\footnote{\textcolor{reviewed_color}{https://github.com/Fields2Cover/fields2cover\_ros}} package provides functions that convert Fields2Cover data types into ROS messages. Services are created to execute modules directly from ROS topics. Launch files are used to script examples of the package. RVIZ-support is also provided to visualize the results of the modules. Methods, objective functions and parameters can be modified on real time thanks to \textit{rqt\_reconfigure}\footnote{http://wiki.ros.org/rqt\_reconfigure}.

\subsection{Design \& Implementation}
\label{subsec:design}

Fields2Cover is implemented using C++17, with a Python interface using Swig \cite{beazley1996swig}, and released under BSD-3 license. The design of Fields2Cover aims to serve both scientists and service providers and is intended to be easily used.

The reason for making Fields2Cover an open-source library is that doing so encourages the development of additional functionality by providing the code to the community. Likewise, Fields2Cover widely employs open-source libraries from third parties to streamline the development process of state-of-the-art algorithms. For scientists, priority is given to a flexible design, which allows to extend or modify existing algorithms. Additionally, a benchmark against which to compare new solutions is added. For service providers, utility concerns the ability to plan the best coverage path for a given objective function in a straightforward manner. The modularity of Fields2Cover is key to ensure its usefulness for both cases. In addition, the library provides tests, tutorials, and extended documentation\footnote{https://fields2cover.github.io/} to reduce the learning curve.

\section{Results}

\begin{figure}[]
\centering
\includegraphics[width=0.9\linewidth]{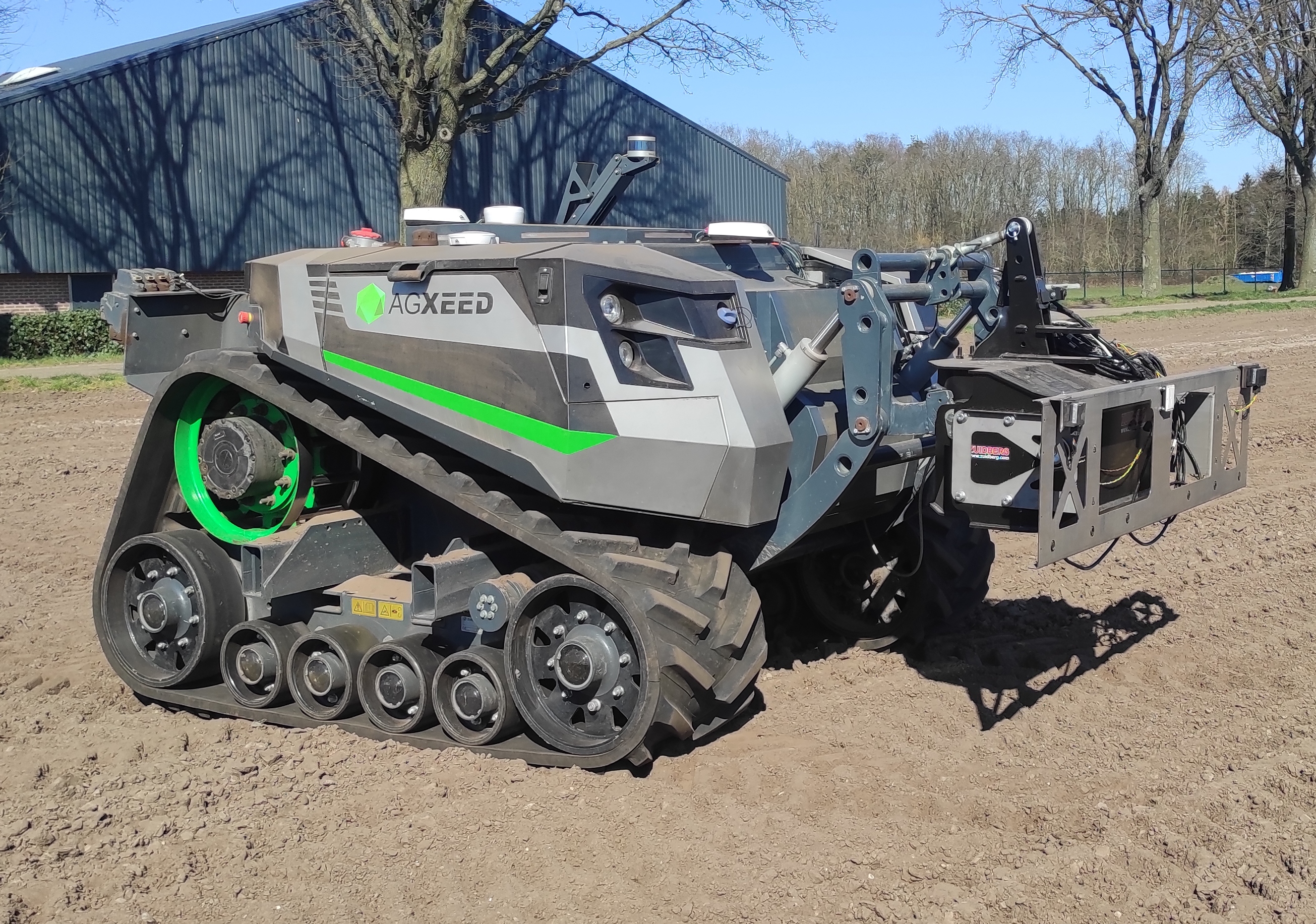}
\caption{The AgBot 5.115T2, from the company Agreed B.V (The Netherlands), is a differential robot with continuous treads. The weight of the robot is 7.8t, the total width of the robot is 2.5 m, the minimum turning radius avoiding excessive soil damage is 2.1 m. For the experiments the operational width of the robot (width of the coupled tool) was assigned the same value as the width of the robot. The AgBot 5.115T2 has 4-cylinder Deutz Diesel Engine, stage 5 with 156hp, and an electric drive train with a maximum speed of 13.5km/h. Some onboard sensors are 2 cameras, a RTK-GNSS receiver and an IMU.}
\label{fig:agbot}
\end{figure}

Several experiments were conducted to demonstrate the functionalities of Fields2Cover. Firstly, coverage paths were created for convex fields from the Nilsson's benchmark \cite{nilsson2020method}. In these simulations, the experiments focus on the optimization of the objective functions and the computation time of those methods. Secondly, real field experiments were conducted in an agricultural field with a commercial robot (Fig. \ref{fig:agbot}) of the company AgXeed B.V (The Netherlands). The aim of the experiment was to program the coverage trajectory of the robot using the Fields2Cover library and assess whether a designed coverage path is efficiently traversed by the robot. The planned path is previously transferred to the robot with Protobuf\cite{Protobuf}. The protobuf message defines the path as timestamps, positions, velocities and orientations. It also contains the geometry of the field boundary to prevent the vehicle from leaving the field. The sensor data collected during the coverage path, such as the \textcolor{reviewed_color}{GNSS} position and the velocity, is returned from the AgBot as a rosbag\cite{Rosbag}.

Experiments were done with a laptop MSI GF627RE with Intel(R) Core(TM) i7-7700HQ CPU @ 2.80GHz (4 cores, 8 threads) with Ubuntu 20.04.5.

\subsection{Simulation results}

Three simulation experiments were performed. Firstly, the optimal route was computed for three different fields to visually inspect the effects of the objective function. Secondly, the coverage path was computed for 38 convex fields with every possible combination of the algorithms provided by the library. The combination of algorithms for creating a coverage path were compared using the path length as the objective function. Thirdly, the time for computing coverage paths was recorded using several objective functions of the Swath Generator module. The relationship between the area of the field and the computation time was found.

The first decision for coverage path planning of a field is the objective function to be optimised by the swath generator (Brute force algorithm). The optimal pitch angle of the swaths may vary with the chosen objective function. Therefore, the first experiment provides examples of optimal swaths for the fields \textit{REC\_A}, \textit{CIR\_B} and \textit{SAL\_B} from the Nilsson's benchmark \cite{nilsson2020method}, which are shown in table \ref{table:covered_fields_obj_comp}. The fields were re-scaled to an area of $100m^2$. If the number of swaths is minimized, the number of turns is also reduced. For instance, fields \textit{CIR\_B} and \textit{SAL\_B} are covered using a single turn. If maximum field coverage is to be achieved, \textit{CIR\_B} needs seven turns while \textit{SAL\_B} needs five. Field coverage is typically achieved when swaths are parallel or perpendicular to one of the edges. In contrast, the swath-length objective function may produce many short swaths (bottom-left of \textit{CIR\_B} with swath length), that reduce the total length of the swaths.

\begin{table}[]
\centering
\caption{Comparison of swaths generated using brute force optimizing one of the three objective functions: sum of swath lengths (minimization problem), number of swaths (minimization problem) and field coverage (maximization problem). The parallel lines inside the field are the centers of the generated swaths.}
\begin{tabular}{| P{1.4cm} || P{1.8cm} | P{1.8cm} | P{1.8cm} |} 
 \hline
Field Name & Swath length & Number Swaths & Field Coverage \\\hline\hline
{REC\_A} & 
\begin{minipage}{1.8cm} \vspace{0.16cm}\includegraphics[width=\linewidth]{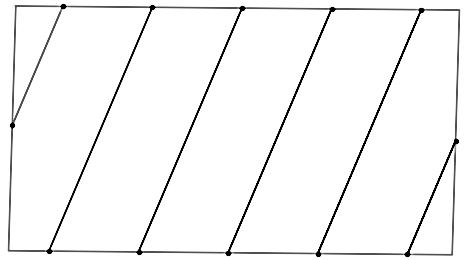}\vspace{0.16cm}\end{minipage} & 
\begin{minipage}{1.8cm} \includegraphics[width=\linewidth]{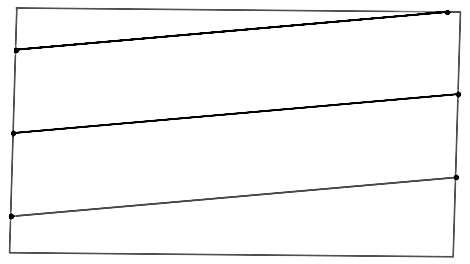}\end{minipage} &
\begin{minipage}{1.8cm}\includegraphics[width=\linewidth]{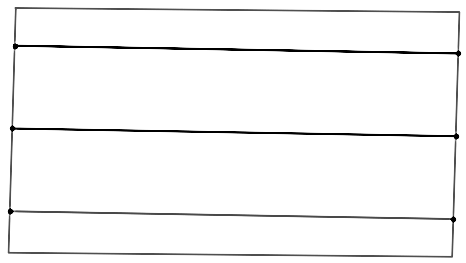}\end{minipage} \\\hline

{CIR\_B} & 
\begin{minipage}{1.8cm} \vspace{0.16cm}\includegraphics[width=\linewidth]{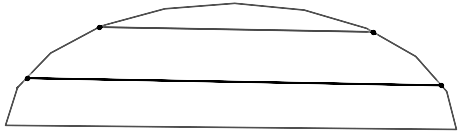}\vspace{0.16cm}\end{minipage} &
\begin{minipage}{1.8cm}\includegraphics[width=\linewidth]{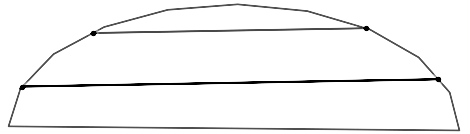}\end{minipage} &
\begin{minipage}{1.8cm}\includegraphics[width=\linewidth]{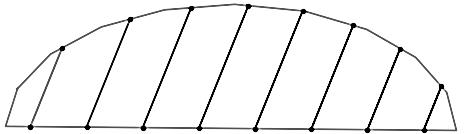}\end{minipage} \\\hline

{SAL\_B} & 
\begin{minipage}{1.8cm} \vspace{0.16cm}\includegraphics[width=\linewidth]{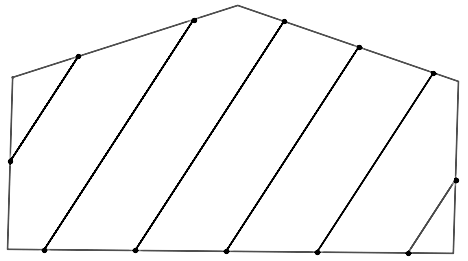}\vspace{0.16cm}\end{minipage} &
\begin{minipage}{1.8cm}\includegraphics[width=\linewidth]{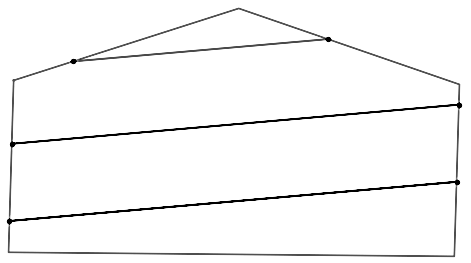}\end{minipage} &
\begin{minipage}{1.8cm}\includegraphics[width=\linewidth]{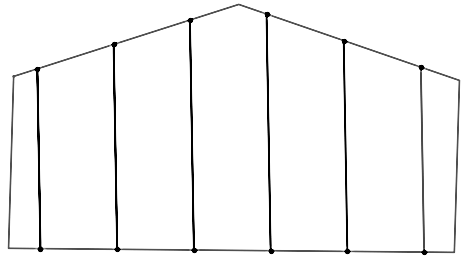}\end{minipage} \\\hline
\end{tabular}
\label{table:covered_fields_obj_comp}
\end{table}    
    
\begin{figure*}[]
\centering
\includegraphics[width=.85\linewidth]{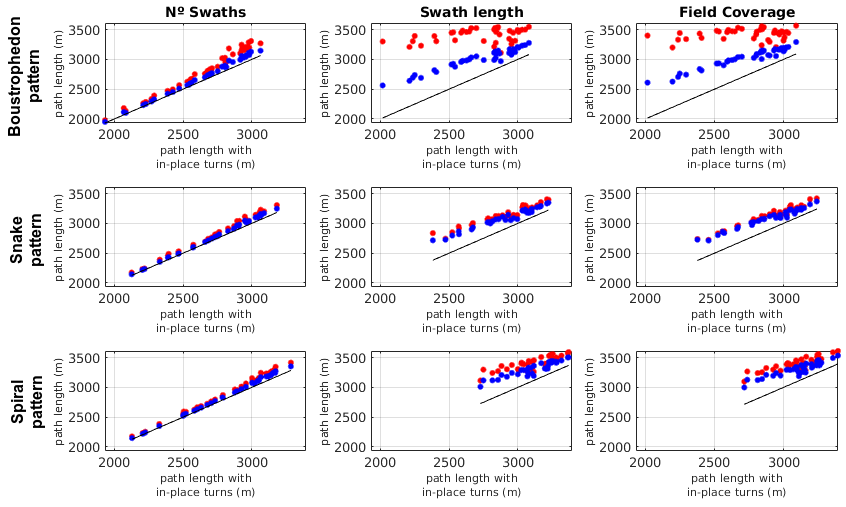}
\caption{Coverage path length comparison. Columns refer to the objective function that was optimized using the brute force algorithm, while each row refers to a particular route planner pattern. \textcolor{reviewed_color}{Each subplot represents the optimized path length (y-axis) with respect to the optimised path length for zero-radius turns (x-axis). A single dot represents a coverage path length for a field with a chosen size from the Nilsson's benchmark \cite{nilsson2020method}. Red dots correspond to paths using Dubins' curves, blue dots are for Reeds-Shepp's curves. The black lines show the 1:1 relations.}}
\label{fig:costs_path}
\end{figure*}    
    
The second experiment was conducted using 38 convex fields of the Nilsson's benchmark \cite{nilsson2020method}, re-scalated to an area of $1$ ha (Fig. \ref{fig:costs_path}). For each field, a headland of $7.5m$ (three times the operational width of the robot) was generated with the constant width generator. Next, the brute force algorithm was used to generate the optimal swaths for each objective function criterion shown in Table \ref{table:covered_fields_obj_comp}. The route planners sorted the swaths with the boustrophedon, snake or spiral (bulk of 6 swaths) pattern. \textcolor{reviewed_color}{Lastly, the computed path length ($L_R$) was used for comparing the coverage paths computed with Dubins' and Reeds-Shepp's curves against the length of paths with in-place turns ($L_0$), which have the least possible path length for a holonomic vehicle. Each column of Fig. \ref{fig:costs_path} refers to the objective function criterion that was optimized by the swath generator and each row denotes a particular route planner pattern. A subplot represents the optimized path length (y-axis) with respect to the optimised path length for in-place turns (x-axis). A single dot represents a computed coverage path, with position ($L_0^i$, $L_R^i$) for the $i^th$ coverage path. The color of the dots denotes the type of curve (Dubin or Reed-Shepp), while black lines represent the 1:1 relation. Greater values for $L_0$ imply that the route generated is longer and distance between swaths is larger. The difference between $L_R^i$ and $L_0^i$ manifests the length of the turns. This difference relates to the time that the machinery is non-productive. Therefore, a substantial difference between the black line and the coloured dots denotes a path for which turning takes more time}. As shown in Fig. \ref{fig:costs_path}, a percentage between $0.5\%$ and $50\%$ of the coverage path was spent on turns. When the number of turns is reduced, the distance traveled is reduced accordingly. The distance used for turning increases when the boustrophedon pattern is applied since a shorter width between swaths requires a larger turn to comply with the minimum turning radius requirement. For instance, in the first column of the figure \ref{fig:costs_path}, the difference between the path length using Dubins' curves and in-place turns is smaller than in the other columns. Field coverage and swath length behaved equally in terms of coverage path length. With any of the objective functions presented, the boustrophedon pattern produced the shortest pattern with in-place turns, the snake pattern was the second and the spiral pattern the longest. The length of the boustrophedon pattern increases when the minimum turning radius is required.

\begin{figure}[t!]
\centering
\includegraphics[width=\linewidth]{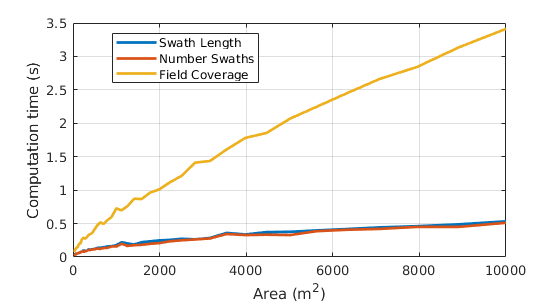}
\caption{Time required to compute a path according to the objective function used. Algorithms used are constant width headland generator, parallel brute force for swath generation, Boustrophedon route order and Dubins' curves.}
\label{fig:time_area}
\end{figure}

In the last simulated experiment, the computation time of planning a coverage path was measured in relation to the area of the field and the objective function of the swath generator (Fig. \ref{fig:time_area}). The constant headland width was set to three times the width of the robot. Next, the parallel brute force algorithm optimized the pitch angle of the swaths, which were sorted using a boustrophedon pattern. Finally, the path planner used Dubins' curves to create the coverage path. \textcolor{reviewed_color}{This experiment measures the computation time for coverage paths according to the three swath generator's objective functions in relation to the area of the field}. Fields2Cover computed a coverage path for a field of $1$ ha in less than 3.5 seconds using Field coverage as the objective function, while only 0.5 seconds were needed using the number of swaths or the swath length as the objective functions. Since the computation of \textcolor{reviewed_color}{the number of swath and swath length} is proportional to the number of swaths and the number of swaths is proportional to the width of the field perpendicular to the driving direction, the computational time grows proportional to the square root of the area of the field. The computation time \textcolor{reviewed_color}{using the latter two} objective functions can be approximated by:
\begin{equation}
    T_c = C_0 * \frac{\sqrt{A_{\bar{hl}}}}{R_w} + C_1
\end{equation}
where $T_c$ is the computation time, $C_0$ and $C_1$ are constants, $A_{\bar{hl}}$ is the area of the field, and $R_w$ is the operational width of the robot. This relationship is only true when the field is convex so it can be covered with \textcolor{reviewed_color}{a single} pattern. 

The field coverage is computationally the most demanding objective function because it computes the difference between the field and the union of the areas of each swath. \textcolor{reviewed_color}{The computation time of this objective function grows linearly with the area of the field}. Geometrical operations such as \textit{'difference'} and \textit{'union'} are more expensive than returning the number of swaths, which is the size of the vector of swaths. 

Computational time analysis focused on the objective function of the brute force algorithm which consumes more than 80\% of the total time of the coverage path planning.

\subsection{Field experiment}

\begin{figure}[]
\centering
\includegraphics[width=\linewidth]{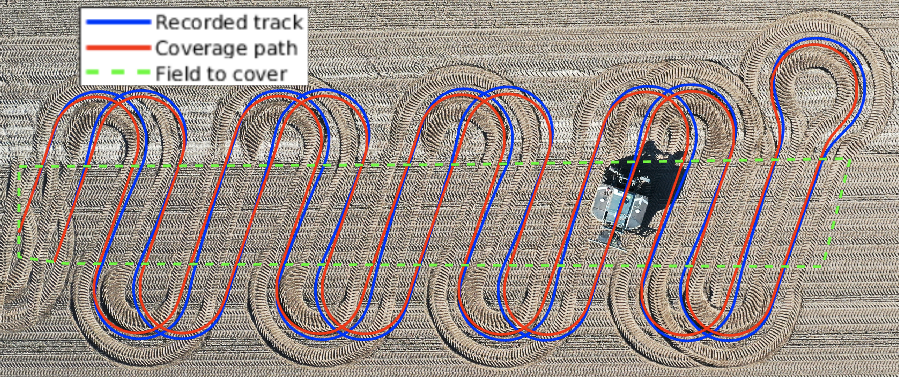}
\caption{AgBot covering a narrow area (shape on green). The coverage path plan in red and the position of the AgBot in blue. The AgBot is halfway the coverage task. The starting point is near the left edge of the area.}
\label{fig:drone_curves}
\end{figure}

A field experiment was conducted using the AgBot shown in (Fig. \ref{fig:agbot}). In the extreme case shown in Figure \ref{fig:drone_curves}, the AgBot covered an elongated narrow area. Objective functions like the minimum swath length or the number of turns would produce swaths parallel to the longest edge of the field. However, here we show a coverage pattern given a custom angle that allows observing the turns in the field. The produced swaths were sorted using the Snake pattern and connected by Dubins' curves.
The difference between the planned path and the recorded track in Figure \ref{fig:drone_curves} \textcolor{reviewed_color}{can be attributed to the planned minimum turning radius being shorter than permissible for the AgBot. Therefore, the recorded GNSS data show slightly wider turns than the planned path}. 
Turns made with the snake pattern always skip one swath, except for the turn at the rightmost part of the field where the coverage direction changed. This turn is sharper, causing wider tracks on the ground, greater soil slippage, and thus more soil damage \cite{janulevivcius2009slippage}. Despite the slippage, the AgBot was capable of covering the field with the path designed by the library routines.

\section{Conclusions \& Future work} 
In this work, we introduced Fields2Cover, a Coverage Path Planning open-source library for agricultural vehicles. Fields2Cover was implemented to bundle the research knowledge on this topic and to help other developers to accelerate their projects. Currently, it supports the creation of coverage paths for convex fields, with a flexible and simple structure thanks to its modular design. The library has four modules, which are: the headlands generator, with a constant width headlands generator; the swath generator, with a brute force optimizer; the route planner, with three types of patterns; and the path planner, with Dubins' and Reed-Shepp's curves. The last three modules have their own objective functions specific to their domains. Fields2Cover was tested using simulation with a public benchmark and in a real field.

Fields2Cover is an on-going project, which means the functionality of the library will be expanded in the coming years. Future developments are supported and maintained by the first author of this paper, with the collaboration of the open-source community. \textcolor{reviewed_color}{Assumptions, such as flat topography, convex fields, enough maneuvering space in the headlands, absence of capacity limits and planning for arable crops, were made to reduce complexity for the first release. Non-convex fields with obstacles, sloping land, capacitated vehicles and permanent crops such as orchards provide challenges for further research and development, part of which are considered within the Field2Cover project.}

\textcolor{reviewed_color}{Since the release of this software library, a community has formed around the coverage path planning problem. In less than 6 months since its release, many developers have shown their support for this project through assigning 150 github stars, code contributions and by providing suggestions for future versions. This community support shows the value of the project, which emphasizes the relevance of addressing the coverage path planning problem.}

\section*{\small Acknowledgement}
This publication is part of the project "Fields2Cover: Robust and efficient coverage paths for autonomous agricultural vehicles" (with project number ENPPS.LIFT.019.019 of the research programme Science PPP Fund for the top sectors which is (partly) financed by the Dutch Research Council (NWO). 

\bibliographystyle{ieeetr}
\bibliography{references} 

\begin{thebibliography}{10}

\bibitem{christiaensen2020future}
L.~Christiaensen, Z.~Rutledge, and J.~E. Taylor, ``The future of work in
  agriculture: Some reflections,'' {\em World Bank Policy Research Working
  Paper}, no.~9193, 2020.

\bibitem{bormann2018indoor}
R.~Bormann, F.~Jordan, J.~Hampp, and M.~H{\"a}gele, ``Indoor coverage path
  planning: Survey, implementation, analysis,'' in {\em 2018 IEEE International
  Conference on Robotics and Automation (ICRA)}, pp.~1718--1725, IEEE, 2018.
\newblock \url{https://doi.org/10.1109/ICRA.2018.8460566}.

\bibitem{jensen2020near}
K.~R. Jensen-Nau, T.~Hermans, and K.~K. Leang, ``Near-optimal area-coverage
  path planning of energy-constrained aerial robots with application in
  autonomous environmental monitoring,'' {\em IEEE Transactions on Automation
  Science and Engineering}, vol.~18, no.~3, pp.~1453--1468, 2020.
\newblock \url{https://doi.org/10.1109/TASE.2020.3016276}.

\bibitem{hameed2017coverage}
I.~A. Hameed, ``Coverage path planning software for autonomous robotic lawn
  mower using dubins' curve,'' in {\em 2017 IEEE International Conference on
  Real-time Computing and Robotics (RCAR)}, pp.~517--522, IEEE, 2017.
\newblock \url{https://doi.org/10.1109/RCAR.2017.8311915}.

\bibitem{oksanen2009coverage}
T.~Oksanen and A.~Visala, ``Coverage path planning algorithms for agricultural
  field machines,'' {\em Journal of field robotics}, vol.~26, no.~8,
  pp.~651--668, 2009.
\newblock \url{https://doi.org/10.1002/rob.20300}.

\bibitem{RJJxpCoveragePlanninggithub}
R.~Jiaping, ``Rjjxp/coverageplanning github.''
  \url{https://github.com/RJJxp/CoveragePlanning}.
\newblock Accessed: \today.

\bibitem{Nobleogithub}
T.~Clephas, C.~López, and Nobleo, ``Nobleo/full\_coverage\_path\_planner
  github.'' \url{https://github.com/nobleo/full_coverage_path_planner}.
\newblock Accessed: \today.

\bibitem{ipa320ipacoverageplanninggithub}
F.~IPA, ``Ipa320/ipa\_coverage\_planning github.''
  \url{https://github.com/ipa320/ipa_coverage_planning}.
\newblock Accessed: \today.

\bibitem{bormann2016room}
R.~Bormann, F.~Jordan, W.~Li, J.~Hampp, and M.~H{\"a}gele, ``Room segmentation:
  Survey, implementation, and analysis,'' in {\em 2016 IEEE international
  conference on robotics and automation (ICRA)}, pp.~1019--1026, IEEE, 2016.
\newblock \url{https://doi.org/10.1109/ICRA.2016.7487234}.

\bibitem{Ethzaslgithub}
R.~Baehnemann, L.~Liu, and D.~Kleiser, ``Ethz-asl/polygon\_coverage\_planning
  github.'' \url{https://github.com/ethz-asl/polygon_coverage_planning}.
\newblock Accessed: \today.

\bibitem{Ethzaslgithubbahnemann2021revisiting}
R.~B{\"a}hnemann, N.~Lawrance, J.~J. Chung, M.~Pantic, R.~Siegwart, and
  J.~Nieto, ``Revisiting boustrophedon coverage path planning as a generalized
  traveling salesman problem,'' in {\em Field and service robotics},
  pp.~277--290, Springer, 2021.
\newblock \url{https://doi.org/10.1007/978-981-15-9460-1_20}.

\bibitem{irvingvasquezocppgithub}
J.~I. Vasquez, ``Irvingvasquez/ocpp github.''
  \url{https://github.com/irvingvasquez/ocpp}.
\newblock Accessed: \today.

\bibitem{irvingvasquezocppgithubgomez2017optimal}
J.~I.~V. Gomez, M.~M. Melchor, and J.~C.~H. Lozada, ``Optimal coverage path
  planning based on the rotating calipers algorithm,'' in {\em 2017
  International Conference on Mechatronics, Electronics and Automotive
  Engineering (ICMEAE)}, pp.~140--144, IEEE, 2017.
\newblock \url{https://doi.org/10.1109/ICMEAE.2017.11}.

\bibitem{Greenziegithub}
Greenzie, ``Greenzie/boustrophedon\_planner github.''
  \url{https://github.com/Greenzie/boustrophedon_planner}.
\newblock Accessed: \today.

\bibitem{IpianoCoveragePlanninggithub}
A.~Stelter, ``Ipiano/coverage-planning github: Quickopp implementation.''
  \url{https://github.com/Ipiano/coverage-planning}.
\newblock Accessed: \today.

\bibitem{driscoll2011complete}
T.~M. Driscoll, {\em Complete coverage path planning in an agricultural
  environment}.
\newblock PhD thesis, Iowa State University, 2011.

\bibitem{jin2009optimal}
J.~Jin, ``Optimal field coverage path planning on 2d and 3d surfaces,'' {\em
  {}}, 2009.
\newblock \url{https://doi.org/10.31274/etd-180810-3122}.

\bibitem{debruin2014gaos}
S.~de~Bruin, P.~Lerink, I.~J. La~Riviere, and B.~Vanmeulebrouk, ``Systematic
  planning and cultivation of agricultural fields using a geo-spatial arable
  field optimization service: Opportunities and obstacles,'' {\em Biosystems
  Engineering}, vol.~120, pp.~15--24, 2014.
\newblock \url{https://doi.org/10.1016/j.biosystemseng.2013.07.009}.

\bibitem{meuth2008divide}
R.~J. Meuth and D.~C. Wunsch, ``Divide and conquer evolutionary tsp solution
  for vehicle path planning,'' in {\em 2008 IEEE Congress on Evolutionary
  Computation (IEEE World Congress on Computational Intelligence)},
  pp.~676--681, IEEE, 2008.
\newblock \url{https://doi.org/10.1109/CEC.2008.4630868}.

\bibitem{zhou2015quantifying}
K.~Zhou, A.~L. Jensen, D.~D. Bochtis, and C.~G. S{\o}rensen, ``Quantifying the
  benefits of alternative fieldwork patterns in a potato cultivation system,''
  {\em Computers and Electronics in Agriculture}, vol.~119, pp.~228--240, 2015.
\newblock \url{https://doi.org/10.1016/j.compag.2015.10.012}.

\bibitem{spekken2016planning}
M.~Spekken, S.~De~Bruin, J.~P. Molin, and G.~Sparovek, ``Planning machine paths
  and row crop patterns on steep surfaces to minimize soil erosion,'' {\em
  Computers and Electronics in Agriculture}, vol.~124, pp.~194--210, 2016.
\newblock \url{https://doi.org/10.1016/j.compag.2016.03.013}.

\bibitem{dubins1957curves}
L.~E. Dubins, ``On curves of minimal length with a constraint on average
  curvature, and with prescribed initial and terminal positions and tangents,''
  {\em American Journal of mathematics}, vol.~79, no.~3, pp.~497--516, 1957.
\newblock \url{https://doi.org/10.2307/2372560 }.

\bibitem{reeds1990optimal}
J.~Reeds and L.~Shepp, ``Optimal paths for a car that goes both forwards and
  backwards,'' {\em Pacific journal of mathematics}, vol.~145, no.~2,
  pp.~367--393, 1990.
\newblock \url{https://doi.org/10.2140/pjm.1990.145.367}.

\bibitem{backman2015smooth}
J.~Backman, P.~Piirainen, and T.~Oksanen, ``Smooth turning path generation for
  agricultural vehicles in headlands,'' {\em Biosystems Engineering}, vol.~139,
  pp.~76--86, 2015.
\newblock \url{https://doi.org/10.1016/j.biosystemseng.2015.08.005}.

\bibitem{sabelhaus2013using}
D.~Sabelhaus, F.~R{\"o}ben, L.~P.~M. zu~Helligen, and P.~S. Lammers, ``Using
  continuous-curvature paths to generate feasible headland turn manoeuvres,''
  {\em Biosystems engineering}, vol.~116, no.~4, pp.~399--409, 2013.
\newblock \url{https://doi.org/10.1016/j.biosystemseng.2013.08.012}.

\bibitem{hoffmann2022weight}
M.~H{\"o}ffmann, S.~Patel, and C.~B{\"u}skens, ``Weight-optimized nurbs curves:
  Headland paths for nonholonomic field robots,'' in {\em 2022 8th
  International Conference on Automation, Robotics and Applications (ICARA)},
  pp.~81--85, IEEE, 2022.
\newblock \url{https://doi.org/10.1109/ICARA55094.2022.9738525}.

\bibitem{nilsson2020method}
R.~S. Nilsson and K.~Zhou, ``Method and bench-marking framework for coverage
  path planning in arable farming,'' {\em Biosystems Engineering}, vol.~198,
  pp.~248--265, 2020.
\newblock \url{https://doi.org/10.1016/j.biosystemseng.2020.08.007}.

\bibitem{norremark2022field}
M.~N{\o}rremark, R.~S. Nilsson, and C.~A.~G. S{\o}rensen, ``In-field route
  planning optimisation and performance indicators of grain harvest
  operations,'' {\em Agronomy}, vol.~12, no.~5, p.~1151, 2022.
\newblock \url{https://doi.org/10.3390/agronomy12051151}.

\bibitem{TBBsoftware}
Intel, ``oneapi threading building blocks.''
  \url{https://github.com/oneapi-src/oneTBB}.
\newblock Accessed: \today.

\bibitem{beazley1996swig}
D.~M. Beazley {\em et~al.}, ``Swig: An easy to use tool for integrating
  scripting languages with c and c++.,'' in {\em Tcl/Tk Workshop}, vol.~43,
  p.~74, 1996.

\bibitem{Protobuf}
Google, ``Protobuf on c++.''
  \url{https://developers.google.com/protocol-buffers}.
\newblock Accessed: \today.

\bibitem{Rosbag}
T.~Field, J.~Leibs, J.~Bowman, and D.~Thomas, ``Rosbag package.''
  \url{http://wiki.ros.org/rosbag}.
\newblock Accessed: \today.

\bibitem{janulevivcius2009slippage}
A.~Janulevi{\v{c}}ius and K.~Giedra, ``The slippage of the driving wheels of a
  tractor in a cultivated soil and stubble,'' {\em Transport}, vol.~24, no.~1,
  pp.~14--20, 2009.
\newblock \url{https://doi.org/10.3846/1648-4142.2009.24.14-20}.

\end{thebibliography}

\end{document}